\documentclass[11pt,a4paper]{article}
\usepackage[hyperref]{emnlp2018}
\usepackage{times}
\usepackage{latexsym}

\usepackage{url}
\usepackage{amssymb,enumitem}
\usepackage{arydshln}
\usepackage{textcomp}
\usepackage{amsmath,amsfonts,amssymb}
\usepackage{graphicx}
\usepackage{paralist}
\usepackage{subcaption}
\usepackage[hang,flushmargin]{footmisc}
\DeclareCaptionFont{10pt}{\fontsize{10pt}{12pt}\selectfont}
\aclfinalcopy 
\def\aclpaperid{***} 

\renewcommand{\footnotesize}{\fontsize{8pt}{10pt}\selectfont}

\title{Document-Level Neural Machine Translation \\ with Hierarchical Attention Networks}

\author{Lesly Miculicich\textsuperscript{$\dagger$ $\ddagger$} \ \ \  Dhananjay Ram\textsuperscript{$\dagger$ $\ddagger$}  \ \ \  \textbf{Nikolaos Pappas\textsuperscript{$\dagger$ }}  \ \ \  James Henderson\textsuperscript{$\dagger$ }\\ 
	\textsuperscript{$\dagger$ }{Idiap Research Institute, Switzerland}\\
	\textsuperscript{$\ddagger$}{\'{E}cole Polytechnique F\'{e}d\'{e}rale de Lausanne (EPFL), Switzerland}\\
	{\tt \{lmiculicich, dram, npappas, jhenderson\}@idiap.ch}}

\date{}

\begin{document}
	\maketitle

	\begin{abstract}
		Neural Machine Translation (NMT) can be improved by including document-level contextual information. For this purpose, we propose a hierarchical attention model to capture the context in a structured and dynamic manner. The model is integrated in the original NMT architecture as another level of abstraction, conditioning on the NMT model's own previous hidden states. Experiments show that hierarchical attention significantly improves the BLEU score over a strong NMT baseline with the state-of-the-art in context-aware methods, and that both the encoder and decoder benefit from context in complementary ways.
	\end{abstract}
	
	\section{Introduction}\label{sec:intro}
	
	Neural machine translation (NMT) \citep{bahdanau2014neural, wu2016google, vaswani2017attention} trains an encoder-decoder network on sentence pairs to maximize the likelihood of predicting a target-language sentence given the corresponding source-language sentence, without considering the document context.
	By ignoring  discourse connections between sentences and other valuable contextual information, this simplification potentially degrades the coherence and cohesion of a translated document \citep{hardmeier2012discourse, meyer-webber:2013:DiscoMT, simsmith:2017:DiscoMT}.	
	Recent studies \citep{tiedemann-scherrer:2017:DiscoMT, jean2017does,wang-EtAl:2017:EMNLP20179,Zhaopeng2017} have demonstrated that adding contextual information to the NMT model improves the general translation performance, and more importantly, improves the coherence and cohesion of the translated text \citep{bawdenevaluating,lapshinovakoltunski-hardmeier:2017:DiscoMT}. 
	Most of these methods use an additional encoder \citep{jean2017does,  wang-EtAl:2017:EMNLP20179} to extract contextual information from previous source-side sentences. However, this requires additional parameters and it does not exploit the representations already learned by the NMT encoder. 
	More recently, \citet{Zhaopeng2017} have shown that a cache-based memory network 
	performs better than the above encoder-based methods. The cache-based memory keeps past context as a set of words, where each cell corresponds to one unique word keeping the hidden representations learned by the NMT while translating it. 
	However, in this method, the word representations are stored irrespective of the sentences where they occur, and those vector representations are disconnected from the original NMT network. 
	
	We propose to use a hierarchical attention network (HAN) \citep{yang-EtAl:2016:N16-13} to model the contextual information in a structured manner using word-level and sentence-level abstractions. In contrast to the hierarchical recurrent neural network (HRNN) used by \citep{wang-EtAl:2017:EMNLP20179}, here the attention allows dynamic access to the context by selectively focusing on different sentences and words for each predicted word. 
	In addition, we integrate two HANs in the NMT model to account for target and source context.
	The HAN encoder helps in the disambiguation of source-word representations, while the HAN decoder improves the target-side lexical cohesion and coherence. The integration is done by (i) re-using the hidden representations from both the encoder and decoder of previous sentence translations and (ii) providing input to both the encoder and decoder for the current translation. This integration method enables it to jointly optimize for multiple-sentences. 
	Furthermore, we extend the original HAN with a multi-head attention \citep{vaswani2017attention} 
	to capture different types of discourse phenomena. 
	
	Our main contributions are the following:
	\begin{inparaenum}[(i)]
		\item We propose a HAN framework for translation to capture context and inter-sentence connections in a structured and dynamic manner.
		\item We integrate the HAN in a very competitive NMT architecture \citep{vaswani2017attention} and show significant improvement over two strong baselines on multiple data sets.
		\item We perform an ablation study of the contribution of each HAN configuration, showing that contextual information obtained from source and target sides are complementary.
	\end{inparaenum}

	\section{The Proposed Approach}
	The goal of NMT is to maximize the likelihood of a set of sentences in a target language represented as sequences of words $\mathbf{y}=(y_1,...,y_t)$ given a set of input sentences in a source language $\mathbf{x}=(x_1,...,x_m)$ as:
		\vspace{-2mm}
		\begin{equation}
		\max_{\Theta}  \frac{1}{N} \sum_{n=1}^N log(P_{\Theta}(\mathbf{y^n}|\mathbf{x^n}))
		\end{equation}	
	so, the translation of a document $\mathbf{D}$ 
	is made by translating each of its sentences independently. 
	In this study, we introduce dependencies on the previous sentences from the source and target sides:
	\vspace{-5mm}
	\begin{equation}
	\max_{\Theta}  \frac{1}{N} \sum_{n=1}^N  log(P_{\Theta}(\mathbf{y^n}|\mathbf{x^n, D_{x^n}, D_{y^n}}))
	\end{equation}
	where $\mathbf{D_{x^n}}=(\mathbf{x^{n-k},...,x^{n-1}})$ and $\mathbf{D_{y^n}}=(\mathbf{y^{n-k},...,y^{n-1}})$ denote the previous $k$ sentences from source and target sides respectively.
	The contexts $\mathbf{D_{x^n}}$ and $\mathbf{D_{y^n}}$ are modeled with HANs.
	
	\subsection{Hierarchical Attention Network}
	The proposed HAN has two levels of abstraction. The word-level abstraction summarizes information from each previous sentence $j$ into a vector $s^j$ as: 
	\vspace{-5mm}
	\begin{align}
	q_w&=f_w(h_t)\\
	s^j&=\underset{i}{\textrm{MultiHead}}(q_w, h_{i}^j)	
	\end{align}
	\vspace{-6mm}
	
	\noindent
	where $h$ denotes a hidden state of the NMT network. In particular, $h_t$ is the last hidden state of the word to be encoded, or decoded at time step $t$, and $h_i^j$ is the last hidden state of the $i$-th word of the $j$-th sentence of the context. The function $f_w$ is a linear transformation to obtain the \emph{query} $q_w$. We used the $\textrm{MultiHead}$ attention function proposed by \citep{vaswani2017attention} to capture different types of relations among words. It matches the \emph{query} against each of the hidden representations $h_i^j$ (used as \emph{value} and \emph{key} for the attention).
	
	The sentence-level abstraction summarizes the contextual information required at time $t$ in $d_t$ as:
	\vspace{-1mm}
	\begin{align}
	q_s&=f_s(h_t)\\
	d_t&=\textrm{FFN}(\underset{j}{\textrm{MultiHead}}(q_s, s^j))
	\end{align}
	\vspace{-4mm}
	
	\noindent
	where $f_s$ is a linear transformation, $q_s$ is the query for the attention function, $\textrm{FFN}$ is a position-wise feed-forward layer \citep{vaswani2017attention}. Each layer is followed by a normalization layer \citep{lei2016}.

	\begin{figure}[t]
		\centering	
		\includegraphics[width=.9\linewidth]{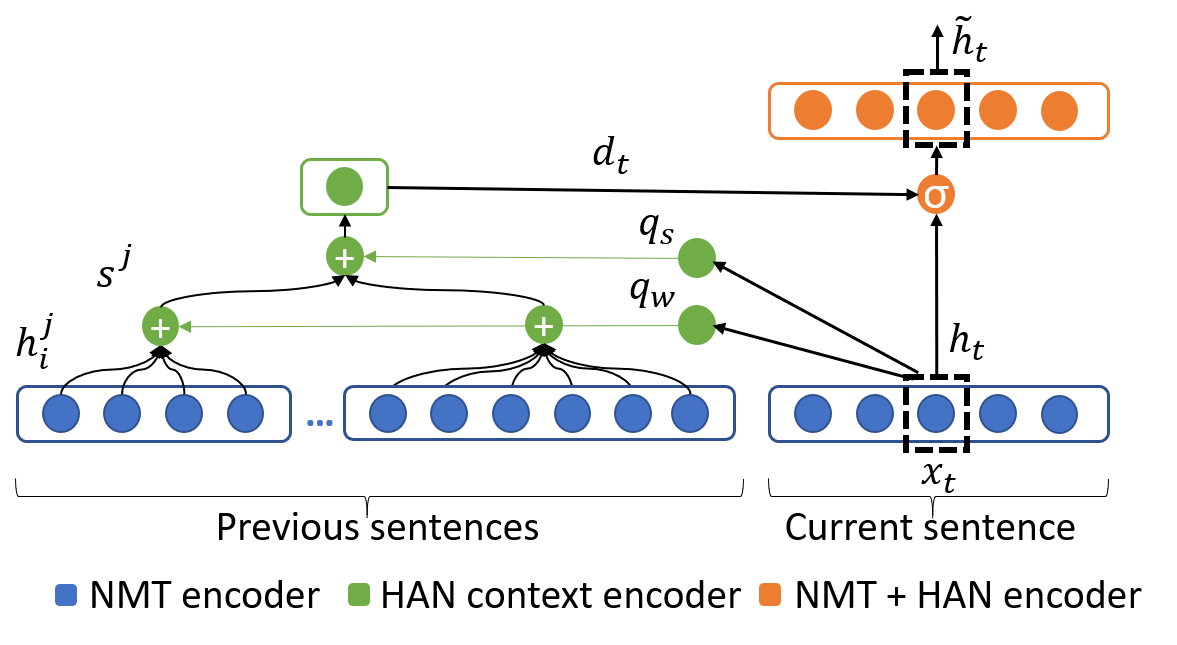}
		\caption{Integration of HAN during encoding at time step $t$, $\tilde{h_t}$ is the context-aware hidden state of the word $x_t$. Similar architecture is used during decoding.}
		\label{fig:HAN}
		\vspace{-2mm}
	\end{figure}

	\subsection{Context Gating}
	We use a gate \citep{Zhaopeng2017, TACL948} to regulate the information at sentence-level $h_t$ and  the contextual information at document-level $d_t$.  The intuition is that different words require different amount of context for translation:
	\vspace{-2mm}
	\begin{align}
	\lambda_t&=\sigma(W_h h_t + W_d d_t)\\
	\widetilde{h}_t &= \lambda_t h_t + (1- \lambda_t)d_t	
	\end{align}	
	\vspace{-4mm}
	
	\noindent
	where $W_h, W_p$ are parameter matrices, and $\widetilde{h}_t$ is the final hidden representation for a word $x_t$ or $y_t$.

	\subsection{Integrated Model}
	The context can be used during encoding or decoding a word, and it can be taken from previously encoded source sentences, 
	previously decoded target sentences, or from previous alignment vectors 
	(i.e. context vectors \citep{bahdanau2014neural}). The different configurations will define the input \emph{query} and \emph{values} of the attention function. In this work we experiment with five of them: one at encoding time, three at decoding time, and one combining both. At encoding time the \emph{query} is a function of the hidden state $h_{x_t}$ of the current word to be encoded $x_t$, and the \emph{values} are the encoded states of previous sentences $h_{x_i}^j$ (HAN encoder). At decoding time, the \emph{query} is a function of the hidden state $h_{y_t}$ of the current word to be decoded $y_t$, and the \emph{values} can be
	\begin{inparaenum}[(a)]
		\item  the encoded states of previous sentences $h_{x_i}^j$ (HAN decoder \emph{source}),
		\item the decoded states of previous sentences $h_{y_i}^j$ (HAN decoder), and 
		\item the alignment vectors $c_{i}^j$ (HAN decoder \emph{alignment}).
	\end{inparaenum}
	Finally, we combine complementary target-source sides of the context by joining HAN encoder and HAN decoder.
	Figure~\ref{fig:HAN} shows the integration of the HAN encoder with the NMT model; a similar architecture is applied to the decoder. The output $\tilde{h}_t$ is used by the NMT model as replacement of $h_t$ during the final classification layer.

	\section{Experimental Setup}
	\subsection{Datasets and Evaluation Metrics}
	We carry out experiments with Chinese-to-English (Zh-En) and Spanish-to-English (Es-En) sets on three different domains: talks, subtitles, and news. 
	
	TED Talks is part of the IWSLT 2014 and 2015 \citep{cettoloEtAl:EAMT2012,cettolo2015iwslt} evaluation campaigns\footnote{\url{https://wit3.fbk.eu}}. We use \emph{dev2010} for development; and \emph{tst2010-2012} (Es-En), \emph{tst2010-2013} (Zh-En) for testing.	
	The Zh-En subtitles corpus is a compilation of TV subtitles designed for research on context \citep{wang2018aaai}. 
	In contrast to the other sets, it has three references to compare. 
	The Es-En corpus is a subset of OpenSubtitles2018 \citep{lison2016opensubtitles2016}\footnote{\url{http://www.opensubtitles.org}}. We randomly select two episodes for development and testing each. 	
	Finally, we use the Es-En News-Commentaries11\footnote{\url{http://opus.nlpl.eu/News-Commentary11.php}} corpus which has document-level delimitation. We evaluate on WMT sets \cite{bojar-EtAl:2013:WMT}: \emph{newstest2008} for development, and \emph{newstest2009-2013} for testing. A similar corpus for Zh-En is too small to be comparable.
	Table~\ref{tab:data} shows the corpus statistics.
	
	For evaluation, we use BLEU score \citep{papineni-EtAl:2002:ACL} (\emph{multi-blue}) 
	on \emph{tokenized} text, and we measure significance with the paired bootstrap re-sampling method proposed by \citet{koehn:2004:EMNLP} (implementations by \citet{koehn-EtAl:2007:PosterDemo}).

	\subsection{Model Configuration and Training}
	As baselines, we use a NMT transformer, and a context-aware NMT transformer with cache memory which we implemented for comparison following the best model described by \citet{Zhaopeng2017}, with memory size of 25 words. 
	We used the OpenNMT \citep{opennmt} implementation of the transformer network. The configuration is the same as the model called ``base model'' in the original paper \cite{vaswani2017attention}. The encoder and decoder are composed of 6 hidden layers each. 
	All hidden states have dimension of 512, dropout of 0.1, and 8 heads for the multi-head attention. The target and source vocabulary size is 30K. The optimization and regularization methods were the same as proposed by \citet{vaswani2017attention}.  \
	Inspired by \citet{Zhaopeng2017} we trained the models in two stages. 
	First we optimize the parameters for the NMT without the HAN, then we proceed to optimize the parameters of the whole network. We use $k=3$ previous sentences, which gave the best performance on the development set.

		\begin{table*}[h]
		\center
		\small
		{\def\arraystretch{1.03}\tabcolsep=0.6pt
			\begin{tabular}{l c l c l c l c l c l} \hline
				& \multicolumn{4}{c}{\textbf{TED Talks}} & \multicolumn{4}{c}{\textbf{Subtitles}} & \multicolumn{2}{c}{\textbf{News}} \\
				&\multicolumn{2}{c}{\textbf{Zh\textendash En}} & \multicolumn{2}{c}{\textbf{Es\textendash En}}& \multicolumn{2}{c}{\textbf{Zh\textendash En \footnotemark}}& \multicolumn{2}{c}{\textbf{Es\textendash En}}  & \multicolumn{2}{c}{\textbf{Es\textendash En}} \\   
				\textbf{Models} &  \multicolumn{1}{c}{BLEU} &  \multicolumn{1}{c}{$\Delta$} & \multicolumn{1}{c}{BLEU} &  \multicolumn{1}{c}{$\Delta$} & \multicolumn{1}{c}{BLEU} &  \multicolumn{1}{c}{$\Delta$} & \multicolumn{1}{c}{BLEU} &  \multicolumn{1}{c}{$\Delta$} &\multicolumn{1}{c}{BLEU} &  \multicolumn{1}{c}{$\Delta$}  \\ \hline
				NMT transformer & 16.87 & & 35.44 && 28.60 && 35.20&& 21.36 & \\ 
				+ cache \citep{Zhaopeng2017} & 17.32 & $(+0.45)^{***}$  & 36.46 & $(+1.02)^{***}$ &  28.86 & $(+0.26)$ &  35.49 & $(+0.29)$  &  22.36 & $(+1.00)^{***}$\\ \hdashline
				+ HAN encoder & 17.61 & ${(+0.74)}^{***}_{\dag\dag}$ & 36.91 & $(+1.47)^{***}_{\dag\dag}$ &  29.35 & $(+0.75)^{*}_{\dag}$ &  35.96  & $(+0.76)^{*}_\dag$  & 22.36 & $(+1.00)^{***}$  \\ 
				+ HAN decoder &  17.39 & $(+0.52)^{***}$ & 37.01 & $(+1.57)^{***}_{\dag\dag\dag}$	 & 29.21 & $(+0.61)^{*}$  & 35.50 & $(+0.30)$ &  22.62 & $(+1.26)^{***}_{\dag\dag\dag}$ \\
				+ HAN decoder \emph{source}  &  17.56 & $(+0.69)^{***}_{\dag\dag}$ & 36.94  & $(+1.50)^{***}_{\dag\dag}$	 & 28.92 & $(+0.32)$  & 35.71 & $(+0.51)^*$ &  22.68 & $(+1.32)^{***}_{\dag\dag\dag}$\\
				+ HAN decoder  \emph{alignment} &  17.48 & $(+0.61)^{***}_{\dag}$ & 37.03 & $(+1.60)^{***}_{\dag\dag\dag}$	 & 28.87 & $(+0.27)$  & 35.63 & $(+0.43)$ &  22.59 & $(+1.23)^{***}_{\dag\dag\dag}$\\
				+ HAN encoder + HAN decoder &  \textbf{17.79} & $(+0.92)^{***}_{\dag\dag\dag}$  & \textbf{37.24} & $(+1.80)^{***}_{\dag\dag\dag}$ & \textbf{29.67} & $(+1.07)^{**}_{\dag}$ &  \textbf{36.23} & $(+1.03)^{**}_{\dag\dag}$ & \textbf{ 22.76} & $(+1.40)^{***}_{\dag\dag\dag}$ \\ \hline
		\end{tabular}}
		\vspace{-2mm}
		\caption{BLEU score for the different configurations of the HAN model, and two baselines. The highest score per dataset is marked in bold. $\Delta$ denotes the difference in BLEU score with respect of the NMT transformer. 
			The significance values with respect to the NMT and the cache method are denoted by $*$, and $\dag$ respectively. The repetitions correspond to the p-values: $^{*}_{\dag}<.05, ^{**}_{\dag\dag}<.01, ^{***}_{\dag\dag\dag}<.001$.}
		\label{tab:bleu}
			\vspace{-2mm}	
	\end{table*}

	\begin{table}
		\center
		\small
		{\def\arraystretch{1.}\tabcolsep=4pt
			\begin{tabular}{l c c c c c} \hline
				& \multicolumn{2}{c}{\textbf{TED Talks}} & \multicolumn{2}{c}{\textbf{Subtitles}} & \multicolumn{1}{c}{\textbf{News}} \\ 
				&Zh\textendash En & Es\textendash En&Zh\textendash En& Es\textendash En & Es\textendash En \\ \hline			
				Training & 0.2M & 0.2M & 2.2M & 4.0M & 0.2M \\
				Development & 0.8K &0.8K & 1.1K & 1.0K & 1.9K\\
				Test & 5.5K & 4.7K & 1.2K& 1.0K& 13.5K \\\hline
		\end{tabular}    }
		\vspace{-2mm}
		\caption{Dataset statistics in \# sentence pairs.}
		\label{tab:data}
		\vspace{-4mm}
	\end{table}

\section{Experimental Results} 
	\subsection{Translation Performance}

	Table~\ref{tab:bleu} shows the BLEU scores for different models. 
	The baseline NMT transformer already has better performance than previously published results on these datasets, and we replicate previous previous improvements from the cache method over the this stronger baseline. All of our proposed HAN models perform at least as well as the cache method.  The best scores are obtained by the combined encoder and decoder HAN model, which is significantly better than the cache method on all datasets without compromising training speed (2.3K vs 2.6K tok/sec).  An important portion of the improvement comes from the HAN encoder, which can be attributed to the fact that the source-side always contains correct information, while the target-side may contain erroneous predictions at testing time.  But combining HAN decoder with HAN encoder further improves translation performance, showing that they contribute complementary information.  The three ways of incorporating information into the decoder all perform similarly. 

	Table~\ref{tab:size} shows the performance of our best HAN model with a varying number $k$ of previous sentences in the test-set.  We can see that the best performance for TED talks and news is archived with 3, while for subtitles it is similar between 3 and 7.

	\subsection{Accuracy of Pronoun/Noun Translations}\label{sec:apt}
	We evaluate coreference and anaphora using the reference-based metric: accuracy of pronoun translation \citep{miculicichwerlen-popescubelis:2017:DiscoMT}, which can be extended for nouns. The list of evaluated pronouns is predefined in the metric, 
	while the list of nouns was extracted using NLTK POS tagging  \citep{P06-4018}.  The upper part of Table~\ref{tab:disc} shows the results. For nouns, the joint HAN achieves the best accuracy with a significant improvement compared to other models, showing  that target and source contextual information are complementary.  
	Similarity for pronouns, the joint model has the best result for TED talks and news. However, HAN encoder alone is better in the case of subtitles. Here HAN decoder produces mistakes by repeating past translated personal pronouns. Subtitles is a challenging corpus for personal pronoun disambiguation because it usually involves dialogue between multiple speakers.

	\subsection{Cohesion and Coherence Evaluation}\label{sec:cohe}
	We use the metric proposed by \citet{wong-kit:2012:EMNLP-CoNLL} to evaluate lexical cohesion. It is defined as the ratio between the number of repeated and lexically similar content words over the total number of content words in a target document. The lexical similarity is obtained using WordNet. Table~\ref{tab:disc} (bottom-left) displays the average ratio per tested document. In some cases, HAN decoder achieves the best score because it produces a larger quantity of repetitions than other models. However, as previously demonstrated in \ref{sec:apt}, repetitions do not always make the translation better. 
	Although HAN boosts lexical cohesion, the scores are still far from the human reference, so there is room for improvement in this aspect.  	
	
	For coherence, we use a metric based on Latent Semantic Analysis (LSA) \citep{foltz1998measurement}. LSA is used to obtain sentence representations, 
	then cosine similarity is calculated from one sentence to the next, and the results are averaged to get a document score. We employed the pre-trained LSA model \emph{Wiki-6} from \citep{L14-1349}. Table~\ref{tab:disc} (bottom-right) shows the average coherence score of documents. The joint HAN model consistently obtains the best coherence score, but close to other HAN models. Most of the improvement comes from the HAN decoder. 
	\footnotetext{\emph{NIST BLEU}: NMT transformer 35.99, cache 36.52, and HAN 37.15.}

	 \begin{table}[t]
		\center
		\small
		{\def\arraystretch{1.}\tabcolsep=4pt
			\begin{tabular}{c c c c c c} \hline
				& \multicolumn{2}{c}{\textbf{TED Talks}} & \multicolumn{2}{c}{\textbf{Subtitles}} &  \multicolumn{1}{c}{\textbf{News}}  \\ 
				\textbf{\textit{k}} &Zh\textendash En & Es\textendash En&Zh\textendash En& Es\textendash En & Es\textendash En  \\ \hline
				
				1 & 17.70 & 37.20 & 29.35 & 36.20 & 22.46 \\
				3 & \textbf{17.79} & \textbf{37.24} & 29.67 & \textbf{36.23} & \textbf{22.76} \\
				5 & 17.49 & 37.11 & \textbf{29.69} & 36.22 &22.54 \\
				7 & 17.00 & 37.22 & 29.64 & 36.21 & 22.64 \\ \hline 
		\end{tabular}}
		\vspace{-2mm}
		\caption{Performance for variable context sizes $k$ with the HAN encoder + HAN decoder. }
		\label{tab:size}
		\vspace{-4mm}
	\end{table}
	
	\begin{table*}[h]
		\center
		\small
		{\def\arraystretch{1.}\tabcolsep=4pt
			\begin{tabular}{l c  c  c  c  c  | c  c  c  c  c } \hline
				& \multicolumn{5}{c|}{\textbf{Noun Translation}} & \multicolumn{5}{c}{\textbf{Pronoun Translation}} \\
				& \multicolumn{2}{c}{\textbf{TED Talks}} & \multicolumn{2}{c}{\textbf{Subtitles}}  & \multicolumn{1}{c|}{\textbf{News}} & \multicolumn{2}{c}{\textbf{TED Talks}} & \multicolumn{2}{c}{\textbf{Subtitles}}  & \multicolumn{1}{c}{\textbf{News}}\\ 
				\textbf{Model} &Zh\textendash En & Es\textendash En&Zh\textendash En& Es\textendash En &Zh\textendash En & Es\textendash En&Zh\textendash En& Es\textendash En	&Zh\textendash En & Es\textendash En \\ \hline				
				NMT Transformer & 40.16 & 65.97 & 46.65 & 61.79 & 47.94 & 63.44 & 68.00 & 69.71 & 65.83 & 47.22 \\
				+ cache & 40.87 &  66.75 & 46.00 &  61.87 & 49.91 & 63.53 & 68.66 & 69.97 &  66.27 & 49.34 \\\hdashline
				+ HAN encoder & 41.93 & 67.75  & 46.78 & 61.52  & 50.06 & 64.05 & 69.17  & \textbf{71.04} &  \textbf{68.56} & 49.57 \\
				+ HAN decoder & 41.61 & 67.35  & 46.78 & 61.99  & 50.03 & 64.02 & 69.36  & 70.50 & 67.03 & 49.33 \\
				+ HAN encoder + HAN decoder & \textbf{42.99} & \textbf{67.81}  & \textbf{47.43} &  \textbf{62.30}  & \textbf{50.40}  & \textbf{64.35} & \textbf{69.60}  & 70.60 &  67.47 &  \textbf{49.59} \\\hline \hline

				& \multicolumn{5}{c|}{\textbf{Lexical cohesion}} & \multicolumn{5}{c}{\textbf{Coherence}} \\

				NMT Transformer & 54.26 & 51.98  & 51.87 &51.77 & 30.06  &0.298 & 0.299  & 0.283 & 0.262 & 0.279\\\hdashline 
				+ HAN encoder & 54.87 & 52.35  & 51.89 & 52.33 & 30.34 &0.304 &0.299  & 0.285 & 0.262 & 0.280\\
				+ HAN decoder & 54.95 & \textbf{52.43} &  \textbf{52.33}& 52.43 & 30.41 &0.302 &  0.301  &  0.287 &  0.265 & 0.282\\
				+ HAN enc. + HAN dec. &   \textbf{55.40} &52.36 & 51.94 &  \textbf{52.75}&\textbf{30.58} & \textbf{0.305} & \textbf{0.302} &  \textbf{0.287}&   \textbf{0.265} &   \textbf{0.282} \\  \hdashline 
				Human reference & 56.08 & 57.02 & 54.81 & 58.19 & 35.12 &0.310 & 0.314 & 0.296 & 0.270 & 0.298\\ \hline 
		\end{tabular}}
		\caption{Evaluation on discourse phenomena. Noun and pronoun translation: Accuracy with respect to a human reference. Lexical cohesion: Ratio of repeated and lexically similar words over the number of content words. Coherence: Average cosine similarity of consecutive sentences (i.e. average of LSA word-vectors)
		}
		\label{tab:disc}
		\vspace{-3mm}
	\end{table*}

		\begin{table}[t]
		\centering
		\small
		{\def\arraystretch{1.}\tabcolsep=2pt
			\begin{tabular}{|l p{0.85\linewidth}|}
				\multicolumn{2}{c}{ Currently Translated Sentence}\\[0.4em] \hline

				Src.:& y esto es un escape de \textbf{su} estado atormentado . \\
				Ref.:& and that is an escape from \textbf{his} tormented state .\\
				Base:& and this is an escape from \textbf{\textit{its}} $<unk>$ state .\\
				Cache:& and this is an escape from \textbf{\textit{their}} state .\\
				HAN:& and this is an escape from \textbf{his} $<unk>$ state . \\\hline
				
			 \end{tabular}}
		{\def\arraystretch{1.2}\tabcolsep=2pt	
			 \begin{tabular}{|l|}
			\multicolumn{1}{c}{ Context from Previous Sentences} \\ \hline
			HAN decoder context with target. \emph{Query}: \textbf{his} (En)\\	
			\includegraphics[width=0.95\linewidth]{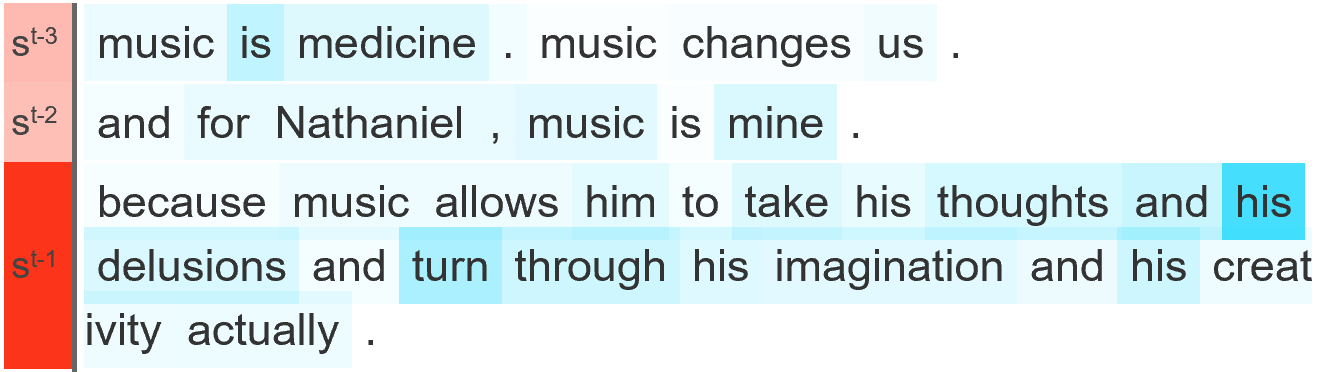}\\ \hline
			HAN encoder context with source. \emph{Query}: \textbf{su} (Es)\\
			\includegraphics[width=0.95\linewidth]{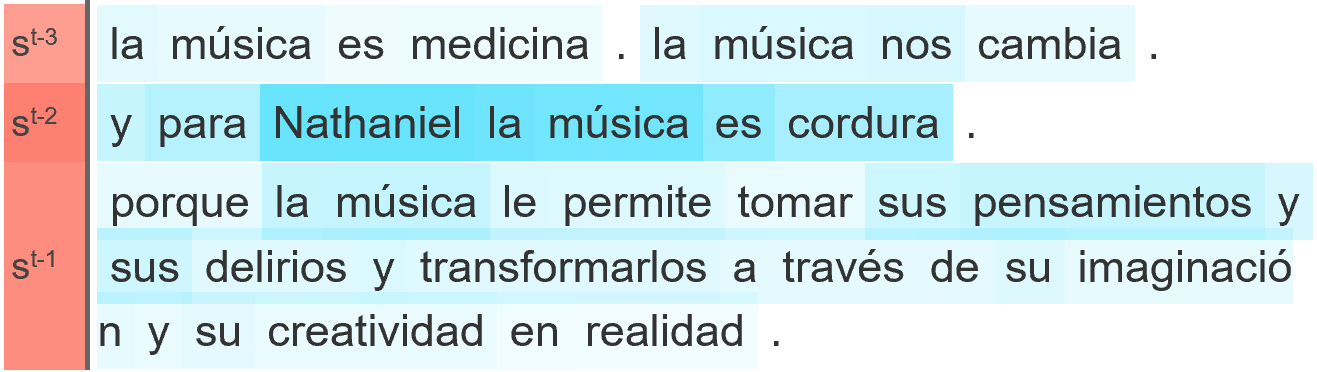} \\ \hline
			\end{tabular}}
		\caption{Example of pronoun disambiguation using HAN (TED Talks Es-En). } 
		\label{tab:sample2}
		\vspace{-6mm}
	\end{table}

	\subsection{Qualitative Analysis}
	Table~\ref{tab:sample2} shows an example where HAN helped to generate the correct translation. The first box shows the current sentence with the analyzed word in bold; and the second, the past context at source and target. For the context visualization  we use the toolkit provided by \citet{pappas-popescubelis:2017:I17-1}.  
	Red corresponds to sentences, and  blue to words. The intensity of color is proportional to the weight.	
	We see that HAN correctly translates  the ambiguous Spanish pronoun ``\emph{su}'' into the English ``\emph{his}''. The HAN decoder highlighted a previous mention of ``\emph{his}'', and the HAN encoder highlighted the antecedent ``\emph{Nathaniel}''. This shows that HAN can capture interpretable inter-sentence connections. More samples with different attention heads are shown in the Appendix~\ref{app}. 

	\section{Related Work}
	\begin{description}[style=unboxed,leftmargin=0cm, itemsep=0pt, parsep=1pt,topsep=0pt, partopsep=0pt]
	 \item[Statistical Machine Translation (SMT)] Initial studies were based on cache memories \citep{tiedemann:2010:DANLP, gong-zhang-zhou:2011:EMNLP}. However, most of the work explicitly models discourse phenomena \citep{simsmith:2017:DiscoMT} 
	such as lexical cohesion \citep{, W12-0117, xiong-EtAl:2013:EMNLP, loaiciga2016predicting, E17-1089, mascarell:2017:DiscoMT}, coherence \citep{born-mesgar-strube:2017:DiscoMT}, and coreference \citep{E17-2104, miculicich:2017:CORBON}. \citet{hardmeier-EtAl:2013:SystemDemo} introduced the document-level SMT paradigm.
	
	 \item[Sentence-level NMT] Initial studies on NMT enhanced the sentence-level context by using memory networks \citep{wang-EtAl:2016:EMNLP20161}, self-attention \cite{N18-1124,zhang-EtAl:2016:EMNLP20162}, and latent variables \cite{yang-EtAl:2017:EACLshort1}.

	 \item[Document-level NMT] \citet{tiedemann-scherrer:2017:DiscoMT} use the concatenation of multiple sentences as NMT's input/output, \citet{jean2017does} add a context encoder for the previous source sentence, \citet{wang-EtAl:2017:EMNLP20179} includes a HRNN to summarize source-side context, and \citet{Zhaopeng2017} use a dynamic cache memory to store representations of previously translated words. 
	 Recently, \citet{bawdenevaluating} proposed test-sets for evaluating discourse in NMT, \citet{P18-1117} shows that context-aware NMT improves the of anaphoric pronouns, and \citet{P18-1118} proposed a document-level NMT using memory-networks. 

	 \end{description}

	\section{Conclusion}
	We proposed a hierarchical multi-head HAN NMT model\footnote{Code available at \url{https://github.com/idiap/HAN_NMT}. Project Towards Document-Level NMT \cite{miculicich2017towards}} to capture inter-sentence connections. We integrated context from source and target sides by directly connecting representations from previous sentence translations into the current sentence translation. 
	The model significantly outperforms two competitive baselines, and the ablation study shows that target and source context is complementary. It also improves lexical cohesion and coherence, and the translation of nouns and pronouns. 
	The qualitative analysis shows that the model is able to identify important previous sentences and words for the correct prediction. 
	In future work, we plan to explicitly model discourse connections with the help of annotated data, which may further improve translation quality. 

	\section*{Acknowledgments}
	We are grateful for the support of the European Union under the Horizon 2020 SUMMA project n. 688139, see \url{www.summa-project.eu}.

	\bibliography{multi_sent}
	\bibliographystyle{acl_natbib_nourl}

\end{document}